\documentclass[conference]{IEEEtran}
\IEEEoverridecommandlockouts
\usepackage{cite}
\usepackage{amsmath,amssymb,amsfonts}
\usepackage{algorithmic}
\usepackage{graphicx}
\usepackage{textcomp}
\usepackage{xcolor}
\usepackage{pifont}
\def\BibTeX{{\rm B\kern-.05em{\sc i\kern-.025em b}\kern-.08em
    T\kern-.1667em\lower.7ex\hbox{E}\kern-.125emX}}
    
\ifCLASSOPTIONcompsoc
\usepackage[caption=false,font=normalsize,labelfont=sf,textfont=sf]{subfig}
\else
\usepackage[caption=false,font=footnotesize]{subfig}
\fi
\begin{document}

\title{Joint Action Language Modelling for Transparent Policy Execution\\
\thanks{The authors gratefully acknowledge funding from the EU and UKRI in the context of Horizon Europe under the MSCA grant agreement No 101072488 (TRAIL). Special thanks also to Yilun Du for publishing the additional short-term language annotations of the Language-Table dataset, and the team at the Computational Shared Facility at the University of Manchester for providing the resources to train our models.}
\thanks{© 2025 IEEE.  Personal use of this material is permitted.  Permission from IEEE must be obtained for all other uses, in any current or future media, including reprinting/republishing this material for advertising or promotional purposes, creating new collective works, for resale or redistribution to servers or lists, or reuse of any copyrighted component of this work in other works.}
}

\author{\IEEEauthorblockN{Theodor Wulff, 
Rahul Singh Maharjan,  
Xinyun Chi and
Angelo Cangelosi}
\IEEEauthorblockA{Department of Computer Science \\
The University of Manchester\\
Manchester, United Kingdom \\
\{theodor.wulff,rahulsingh.maharjan,xinyun.chi,angelo.cangelosi\}@manchester.ac.uk}}

% \author{\IEEEauthorblockN{Theodor Wulff}
% \IEEEauthorblockA{\textit{Department of Computer Science} \\
% \textit{The University of Manchester}\\
% Manchester, United Kingdom \\
% theodor.wulff@manchester.ac.uk}
% \and
% \IEEEauthorblockN{Angelo Cangelosi}
% \IEEEauthorblockA{\textit{Department of Computer Science} \\
% \textit{The University of Manchester}\\
% Manchester, United Kingdom \\
% angelo.cangelosi@manchester.ac.uk}
% }

\maketitle

\begin{abstract}
An agent's intention often remains hidden behind the black-box nature of embodied policies. 
Communication using natural language statements that describe the next action can provide transparency towards the agent's behavior.
We aim to insert transparent behavior directly into the learning process, by transforming the problem of policy learning into a language generation problem and combining it with traditional autoregressive modelling.
The resulting model produces transparent natural language statements followed by tokens representing the specific actions to solve long-horizon tasks in the Language-Table environment. 
Following previous work, the model is able to learn to produce a policy represented by special discretized tokens in an autoregressive manner. 
We place special emphasis on investigating the relationship between predicting actions and producing high-quality language for a transparent agent.
We find that in many cases both the quality of the action trajectory and the transparent statement increase when they are generated simultaneously. 
\end{abstract}

% \begin{abstract}
% This document is a model and instructions for \LaTeX.
% This and the IEEEtran.cls file define the components of your paper [title, text, heads, etc.]. *CRITICAL: Do Not Use Symbols, Special Characters, Footnotes, 
% or Math in Paper Title or Abstract.
% \end{abstract}


% \documentclass{article}
% \usepackage{graphicx} % Required for inserting images
% \usepackage{appendix}

% \title{Joint Action Language Modelling for Transparent Policy Execution}
% \author{Theodor Wulff, Angelo Cangelosi}
% \date{\today}

% \begin{document}

% \maketitle

\section{Introduction}
% Main Target: Increase Robot Transparency
The field of robotics is progressing towards robots with higher degrees of autonomy~\cite{brohan_rt-1_2023}.
Eventually, robots could be able to collaborate with humans after receiving only very little instruction on how to complete a certain task.
However, as a robotic agent's degree of autonomy increases, its actions tend to remain opaque until the moment they are executed~\cite{Sakai2022autonomous}.
This is especially important for robots deployed in real-life scenarios where the interacting human might not have experience with that agent.
In cases where the robot's behavior does not align with the expectations of the human, this can lead to a loss of trust in the robotic agent and, as a consequence, hinder the effectiveness of collaboration~\cite{Gervasio2018ExplanationTA}. 
To avoid these situations and establish a common ground, humans naturally utilize language, among other behaviours, to coordinate tasks and solve problems effectively~\cite{rieser2005implications}.
We hypothesize that autonomous robotic agents should exhibit similar behavior to collaborate in human-robot teams effectively. 

% Problem: Transparency underresearched in current robotics
Current robotics research focuses primarily on the training of agent behavior by evaluating policy execution and success rates within specific environments~\cite{brohan_rt-1_2023,brohan_rt-2_2023,kim24openvla,belkhale_rt-h_2024}. 
Learning to provide transparency at the same time is rarely a consideration for the development of most agentic systems.
Behavior transparency can be achieved by different means, for example, by providing an outline of the action trajectory before execution or by generating a detailed plan of subsequent actions~\cite{schoett2023transparencySurvey}. 
In this work, we chose the domain that is the most commonly used for communication between humans and robots for transparency: natural language~\cite{schoett2023transparencySurvey}.

% Language Capabilities for communication
The success of Foundation Models, which have been trained on a huge amount of data, especially in the domains of language and vision, has led to researchers in robotics adopting such models in their systems~\cite{kim24openvla,brohan_rt-1_2023,brohan_rt-2_2023,black2024pizero,belkhale_rt-h_2024}.
However, they are mostly used either to learn a policy directly, which is no longer transparent, or to map an observation to a lower-level statement, which is then executed by a separately learned policy.
The capabilities of producing language for communication are then considered separately, if at all. 
We argue that for a robotic agent to exhibit transparent behavior, learning to be transparent should be incorporated into the learning process from the beginning.

% Motivation by VLM based vision language action models
In this work, our aim is to increase the transparency of robot behavior by utilizing a single model to generate the next action and simultaneously provide a natural language statement.
Modern Vision-Language Models (VLMs) form a base for effective grounded communication, as they have shown excellent qualities in image understanding and question-answering tasks~\cite{Beyer2024paligemma}. 
We leverage the general language understanding and generation capabilities of the VLM to a) communicate the agent's subsequent action in natural language, and b) utilize specific action tokens to execute the agent policy by turning the problem into a full language learning task.
This also allows us to specify additional contextual information like the robot's state into the query by mapping it to specific tokens or providing a corresponding language utterance.
% Knowing when to produce a new language statement
%In addition to the ability to produce a statement describing the current action intended for transparency, knowing when to provide a user with a new statement is equally important, which is why we introduce an additional query to model the switch between statements.

Our key contributions can be summarized as:
\begin{itemize}
    \item \textbf{Joint Action Language Generation formulated as a transparent language-learning problem.} We transform the problem of learning the agent's policy to steer the actuators into a natural language processing task which inherently produces transparent statements within the same output. 
    Contrary to prior work~\cite{kim24openvla,black2024pizero,brohan_rt-2_2023} that uses VLMs for policy learning, we specifically investigate the interplay between the production of language statements and low-level actions.
    \item \textbf{Actions benefit from transparency.} Our results show that autoregressively generating a transparent language statement alongside action tokens positively impacts the predicted trajectories as well as the language output.
\end{itemize}

The remainder of this paper is organized as follows: 
Section~\ref{sec:related_work} covers current work on Vision-Language-Action Models and transparency in robotics. 
Section~\ref{sec:method} describes the details of our approach. 
In Sections~\ref{sec:experiments} and~\ref{sec:results} we present our experimental setup and results.
The paper concludes with a discussion and brief summary in Sections~\ref{sec:discussion} and~\ref{sec:conclusion}.
 
\section{Related Work}\label{sec:related_work}
\subsection{Vision-Language-Action Models}
Based on their high generalizability, recent works have started to utilize VLMs in the field of robotics.
Applied as an interface for training language-conditioned vision-based policies, they are termed Vision-Language-Action Models (VLA). 
This is accomplished by either defining a textual representation of the actions and treating the problem as a conditional language generation task or using specific policy heads that produce the action.

% BC-Z
% Jang et  al.~\cite{jang_bc-z_2022} imitation learning on broad language-conditioned tasks, also from human demonstration videos. 
% PaLM-E
Driess et al.~\cite{driess_palm-e_2023} embed robot state, pixel- and object-level visual input, and language in a multimodal token sequence, which is processed by a Large Language Model (LLM). 
A separate control policy performs the actions based on the language guidance provided by the model. 
They train their PaLM-E model~\cite{driess_palm-e_2023} on various multimodal tasks to increase the quality of extracted features.
% Octo
Gosh et al.~\cite{octo_model_team_octo_2024} propose to train a transformer-based generalist agent, called Octo, by training on a wide variety of mixed-modality data. 
% Manip-LLM
Li et al.~\cite{li_manipllm_2024} predict coordinates of gripper-related objects in the field of view using specific queries to the VLM.
% Perceiver-Actor
Shridhar et al.~\cite{shridhar_perceiver-actor_2022} utilize 3D Voxels as input and goal specification to the Perceiver-Actor model.
% OpenVLA 
Kim et al.~\cite{kim24openvla} propose OpenVLA, an open-source VLA that generates action tokens using an LLM backbone, which processes language and visual features.
A specific decoder de-tokenizes the output tokens into low-level actions. 
The model is trained on a multitude of embodiments for generalizability across domains~\cite{kim24openvla}.
% Pi Zero
Black et al.~\cite{black2024pizero} utilize a pretrained PaliGemma Vision-Language Model~\cite{Beyer2024paligemma} and attach an \textit{action expert} to it for policy execution.

% RT-1
The RT-1~\cite{brohan_rt-1_2023} model embeds language and visual input into a joint token representation and produces discrete actions using a transformer head.
% RT-2
Its successor RT-2~\cite{brohan_rt-2_2023} utilizes a central LLM, which processes visual and language input to predict action tokens and was trained across Visual Question Answering among other tasks. 
% RT-H
RT-H~\cite{belkhale_rt-h_2024} employs a two-step querying strategy by first letting the model break down the current task into a short-term action which serves as the context for predicting the next action using the same model.
% ERRA
Similarly, Zhao et al.~\cite{zhao_erra_2023} break down an abstract task description into concretely executable actions whose execution is learned separately.

Although many models have been pretrained on language-generation tasks, little to no emphasis is put on utilizing the language-generative capabilities of the model to simultaneously make the actions more transparent.
The aim is to create an improved language-conditioned policy, in contrast to a policy with explicitly high-quality language output.
We aim to move research a step forward towards agents that have inherently learned to behave transparently.

\subsection{Explainability and Transparency}
Explainability and transparency in robotics are related topics, but differ slightly in their goals.
Explainability aims to answer the \textit{why, what and how} of robot behavior while transparency is mostly concerned with the \textit{what and how}, which facilitates inferring the \textit{why} without explicitly providing it~\cite{schoett2023transparencySurvey}.
Although we focus on transparency in this work, techniques that train a model to provide an explanation contain aspects of transparency as well.
Both concepts have in common that they aim to provide answers on the \textit{what} and \textit{how} with respect to the agent's behavior.

% Interactive Explanation Learning
Work in interactive explanation learning~\cite{teso_explanatory2019} aims to improve model-generated explanations with a human-in-the-loop who acts as a critic of the model's explanations, similar to how LLMs are trained with Reinforcement Learning from Human Feedback (RLHF)~\cite{ouyang2022instructgpt}.
% Learning by Self-Explaining
Stammer et al.~\cite{stammer2024learning} automate the learning process by replacing the human with a surrogate model that acts as the critic in the human's place.
% Explaining Autonomy
Alternatively, Duan et al.~\cite{duan2024aha} train their VLM to provide natural language explanations on robot failures, after specifically pretraining on such cases.
%ArXIv only: Sobrin-Hidalgo et al.~\cite{} generate explanations by processing the robot's log files to generate human-understandable outputs regarding the robot's behaviour.
% investigating Transparency Methods in a Robot WordñLearning System and Their Effects on Human Teaching Behaviors

Leveraging existing LLM or VLMs, Chain-of-thought methods can provide transparency by outlining multiple steps that lead to some solution. However, since these are generated, they can be erroneous themselves.
% Chain of thought, tree of thought etc...
The basic idea is to prompt an LLM or VLM to think step-by-step instead of solely producing an answer to a question and was proposed by Wei et al.~\cite{Wei_cot_2022}.
As such, many variations of this process have been proposed that incorporate different modalities into the intermediate step-by-step thought chain~\cite{ni2024subgoal,liu2023minds} or explore different possible paths along consecutive thought chains~\cite{Yao_tree_2023}.
% Autonomous Justification for Enabling Explainable Decision Support in Human-Robot Teaming
% What is on your mind nico

Other methods like Kerzel et al.~\cite{Kerzel2022WhatsOY} equip a robot with transparent behavior using a variety of modalities that highlight different decisions that take place in the system. 
% Transparency often researched in the fields of psychology and HRI
Besides the progress of transparency in machine learning and robotic applications, transparency is an active topic of research in the fields of psychology and human-robot interaction~\cite{schoett2023transparencySurvey}.
Other research investigates the use of different modalities to provide transparency on robot behavior, even though language and speech are among the most prominent choices~\cite{wallkoetter2021review}.

%Other attempts utilize inherently transparent methods due to e.g. behaviour trees as the underlying structure.
\section{Method}\label{sec:method}
\begin{figure*}
    \centering
    \includegraphics[width=\linewidth]{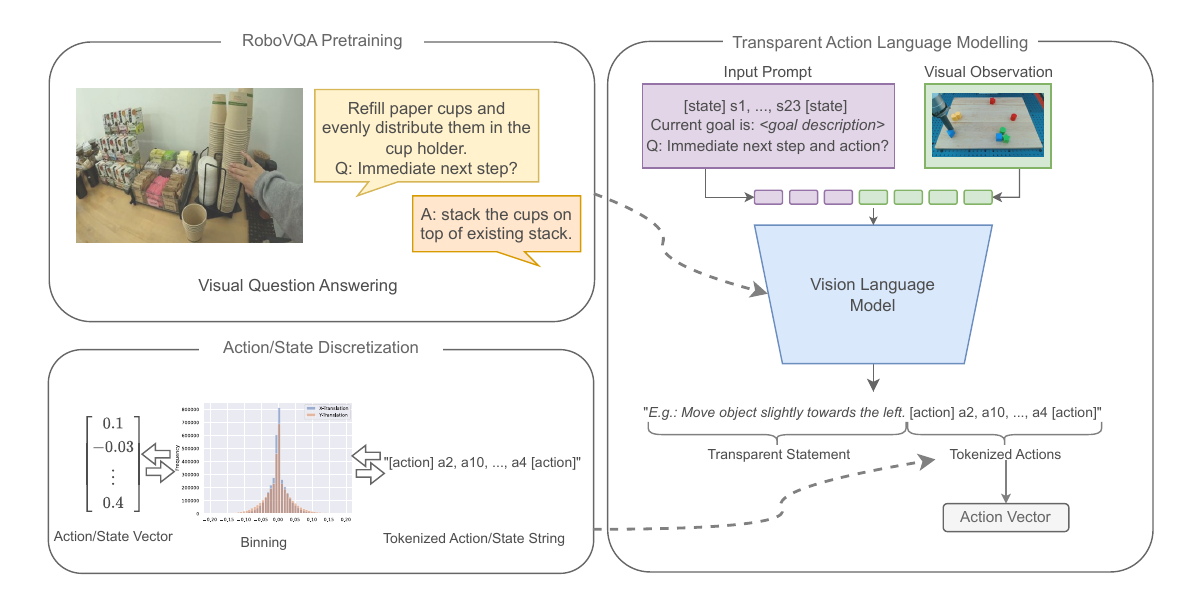}
    \caption{\textbf{Method Overview.} We utilize the Vision-Language Model PaliGemma to produce a transparent statement and action tokens given an input prompt, describing the current task, and the visual observation of the environment. The model is pretrained on visual question answering in robotic settings. We discretize the state and action vectors into special tokens to embed these directly into the input prompt and target strings.}
    \label{fig:method}
\end{figure*}

Our training procedure consists of two main steps. 
First, we pretrain our models on visual question answering to generate transparent statements in robotic settings. 
Then, we train on robotic tasks to generate actions in addition to the transparent statements. 
Figure~\ref{fig:method} visually summarizes our approach.

\subsection{Problem Formulation}
To facilitate the learning process and to show that the model can learn both the transparent statement in conjunction with the policy, we train the model to imitate an expert's behavior and augment its actions with the natural language statement given by the ground truth caption. 
Similar to Jang et al.~\cite{jang_bc-z_2022}, we train our models in a supervised way using language-conditioned behavior cloning.
We extend the problem of language-conditioned behavior cloning~\cite{nair2021learning} with the additional requirement of producing transparent statements.
This problem is often modelled as a partially observable Markov decision process~\cite{puterman1994markov}.
%The problem we investigate can be formulated as a partially observable Markov decision process (POMDP).
Conventially, the goal is to learn to predict the next action that follows an expert's policy given a language instruction and some observation of the environment.
We extend this formulation by providing a long-term language instruction for the input and expanding the output to generate a transparent statement in the form of a short-term natural language description alongside the action.
% Given a long-term language instruction %$l_i \in L$ 
% and an observation, %$o \in O$, where $L$ and $O$ are the sets of possible long-term language instructions and observations respectively
% the goal is to predict the next action %$a \in A$ 
% alongside a transparent statement in the form of a natural language description %$d \in D$ 
% of the performed motion. 

\begin{table}[ht]
    \caption{Prompt Definitions}
    \label{tab:prompts}
    \renewcommand{\arraystretch}{1.3}
    \centering
    \begin{tabular}{c|c}
        Prompt & Target \\\hline
        Next action? & Action tokens \\
        Immediate next step and action?& Description and action tokens \\
%        Update verbal statement? & True/False \\
        Immediate next step? & Description \\\hline
        \hline
        \multicolumn{2}{c}{Context Definitions}\\\hline
        \multicolumn{2}{l}{Current task is: $<$\textit{Instruction}$>$. $<$\textit{Prompt}$>$}\\
        \multicolumn{2}{l}{Given $<$\textit{State}$>$. Current task is: $<$\textit{Instruction}$>$. $<$\textit{Prompt}$>$}
    \end{tabular}
\end{table}
In our case, the observations consist of a camera input from the robot, the current end-effector state, and the language query that contains the long-term goal instruction and the corresponding question.
To embed the action and state vectors into the language prompts, we discretize each dimension of the continuous action/state vector representations by applying a binning strategy and mapping each associated bin onto corresponding special tokens.
Refer to Section~\ref{subsec:discretization} for the detailed tokenization process.

Prior works have not explicitly investigated the quality of statements which have been produced as intermediate output.
While the positive effects of intermediate outputs have been seen, e.g., in Chain-of-Thought-like mechanisms~\cite{Wei_cot_2022,belkhale_rt-h_2024,brohan_rt-2_2023}, the quality of these explanations is not clear. 
Here, we specifically investigate the output quality concerning ground truth statements.

% Difference to explicitly generating a plan: no need to learn the lower-level instruction for each action primitive. We learn both and associate one with the other. Basically one step plan

\subsection{Datasets}
\label{subsec:data}
% The following paragraphs describe the data we used and which additional processing steps we applied.
\subsubsection{RoboVQA}
\label{subsubsec:robovqa}
The RoboVQA~\cite{sermanet2023robovqa} dataset is a recent VQA dataset specific to robotic settings. 
It contains video scenes of different agents operating in robotic scenarios annotated with question-answering pairs.
The questions are asking for different forms of planning, success recognition, and scene understanding.
We utilize the freeform planning questions to pretrain our models. 
Here, the goal is to predict a single natural language statement describing the "immediate next step".
In total, we utilize 94,997 freeform planning question-answer-image triplets to pretrain our models.
% 2742 unique questions, 28852 unique answers

\subsubsection{Language Table}
\label{subsubsec:lang_table}
We split the Language-Table~\cite{lynch2023interactivelanguage} dataset into training, validation and test subsets on an episode level.
An episode consists of a long-horizon goal provided in natural language. 
Furthermore, each episode is split into multiple sub-episodes which are annotated with captions that describe the current \textit{high-level} action. 
A sub-episode ends once the caption changes.
We utilize the captions as our ground truth labels for the agent's transparent statements.
The action trajectory consists of a sequence of \textit{low-level} actions, in this case, 2D vectors referring to the translation of the robot arm's pointer across the board.
This leads to a dataset with 23,019 episodes and 399,846 captions.
We load a batch of episodes and sample $N$ observations for each caption to form our mini-batch before shuffling and passing it into the network. 
To sample the individual frames, we always select the first and last frames and uniformly draw $N-2$ frames from the sequence in between these.
%The last timestep before the caption changes is additionally marked as ("Change natural language statement?", True) while the rest is annotated with False.

% Number of episodes: 23019
% Total number of instructions: 399846
% Train instructions: 320102
% Train unique instructions: 102785
% Val instructions: 39866
% Val unique instructions: 23076
% Test instructions: 39878
% Test unique instructions: 23209
% Test unseen instructions: 8524
% Test Val Difference: 8192
% Val Test Difference: 8235

\subsection{Pretraining}
We pretrain our models on the freeform planning subset of the RoboVQA dataset, which asks the model to predict the subsequent step given the current frame and context.
We hypothesize that this extra fine-tuning step is beneficial since robotic data is sparse and rarely encountered in general web-scraped text-image datasets used to pretrain VLMs~\cite{oneill2024OpenX}.
The answer to the question is a transparent statement: a natural language description of the immediate next action.
Queries are of the form: "Current goal is: \textit{$<$goal description$>$}, immediate next step?"
In some later configurations, we prepend the state information to the query following the tokenization strategy described in Section~\ref{subsec:discretization}.
Refer to Table~\ref{tab:prompts} for an exhaustive overview of prompts used in our setup.

\subsection{Model}
Taking inspiration from Black et al.~\cite{black2024pizero}, we use the PaliGemma model~\cite{Beyer2024paligemma} for our experiments, due to its relatively small size with ca. 3 billion parameters, which reduces training and inference times, and the effective generalizability across different multimodal tasks, which require language grounding in visual inputs.
% It has also been used successfully by Black et al.~\cite{black2024pizero} to learn robotic policies.
% PaliGemma consists of a SigLIP image encoder and a Gemma decoder language model.
We stick to an input resolution of 224x224 pixels for the images, since higher resolutions increase computational requirements, and previous work has not found a worthwhile performance increase.
While we chose the PaliGemma model for the reasons above, our method can be applied to any other Vision-Language Model.
\subsection{Action and State Tokenization}
\label{subsec:discretization}
We limit the range of possible action trajectories to (-0.05, 0.05) along each dimension and state trajectories to (-0.3, 0.35) along the x-axis and (0.2, 0.6) along the y-axis. 
These ranges ensure that we capture at least one standard deviation of each trajectory dimension based on the distribution of our training data. 
We discretize the two-dimensional action and state space by mapping each dimension of the trajectory and robot state to special tokens, which we add to the vocabulary of our model and tokenizer. 
We associate each of the action and state bins with a number to create the special tokens "\textit{a0}" - "\textit{aN}" and "\textit{s0}" - "\textit{sN}" for the tokenized actions and states. 
Additionally, we surround the tokenized state and actions with the special tokens "\textit{[state]"} or "\textit{[action]}" to mark the beginning and end of the state and actions.
We perform the same procedure on the robot state as in RT-2~\cite{brohan_rt-2_2023} and OpenVLA~\cite{kim24openvla}.

% \begin{table}[ht]
%     \caption{Hyperparameters}
%     \label{tab:parameters}
%     \renewcommand{\arraystretch}{1.3}
%     \centering
%     \begin{tabular}{c|c}
%         Parameter & Value \\\hline
%         Frames per episode & 3 \\
%         Learning rate & $10^{-5}$\\
%         Batch size & 256
%     \end{tabular}
% \end{table}

\section{Experiments}\label{sec:experiments}
We train our models on an Nvidia A100 GPU using the Adam optimizer.
The training duration of our models is three epochs, each containing 160,000 examples. 
The mini-batches contain 8 samples, but we accumulate the gradients between mini-batches to reach an effective batch size of 256.
We sample 3 frames from each sub-episode, following the procedure outlined in Section~\ref{subsec:data}, to ensure that our model sees a high variety of captions during training.
% The detailed hyperparameters used for our experiments can be found in Table~\ref{tab:parameters}.
To calculate the MSE and Cosine Similarity between the generated and target trajectories, we map the generated action tokens back to their respective discrete bins and compare the mean value of the bin with the continuous trajectory.
Additionally, we investigate the ability to produce coherent language output using common metrics found in natural language processing: namely the BLEU~\cite{papineni-etal-2002-bleu} and ROGUE-1~\cite{lin-2004-rouge} scores.
All measurements presented in Section~\ref{sec:results} are presented with the corresponding standard deviation across the test set.
%To semantically analyze the alignment of the provided statement and the model we use a pretrained CLIP model to calculate the similarity scores.
% The results of our models on the RoboVQA dataset can be found in Appendix~\ref{app:pretraining}

\section{Results}\label{sec:results}
We provide the results on the language table dataset using supervised imitation learning. 
We investigate both the model's ability to reproduce the expert's action and the quality of the language statements.
Previous work has already shown that VLMs can generate actions for execution on robotic systems~\cite{brohan_rt-1_2023,brohan_rt-2_2023,belkhale_rt-h_2024,kim24openvla}.
Of special interest is the effect that producing the verbal statement alongside a trajectory has on the quality of the trajectories.

\begin{table}[ht]
    \caption{Action Before Language}
    \label{tab:action_first}
    \renewcommand{\arraystretch}{1.3}
    \centering
    \begin{tabular}{c|cccc}        
        % Actions First & ROUGE$\uparrow$ & BLEU$\uparrow$ & CosSim$\uparrow$ & MSE$\downarrow$ \\\hline
        % \ding{55}  & 0.5087$\pm$0.04 & \textbf{0.1511$\pm$0.04} & 0.0106$\pm$0.25 & 0.0024$\pm$0 \\ 
        % \ding{51} & \textbf{0.5247$\pm$0.04} & 0.1334$\pm$0.03 & \textbf{0.0377$\pm$0.26} & 0.0024$\pm$0 \\
        
        Actions First & ROUGE$\uparrow$ & BLEU$\uparrow$ & CosSim$\uparrow$ & MSE$\downarrow$ \\\hline
        \ding{55}  & 0.5087 & \textbf{0.1511} & 0.0106 & 0.0024 \\ 
        \ding{51} & \textbf{0.5247} & 0.1334 & \textbf{0.0377} & 0.0024 \\ 
    \end{tabular}
\end{table}
\begin{figure}[ht]
    \centering
    \subfloat[Effects on language modeling.]{\includegraphics[width=.45\linewidth]{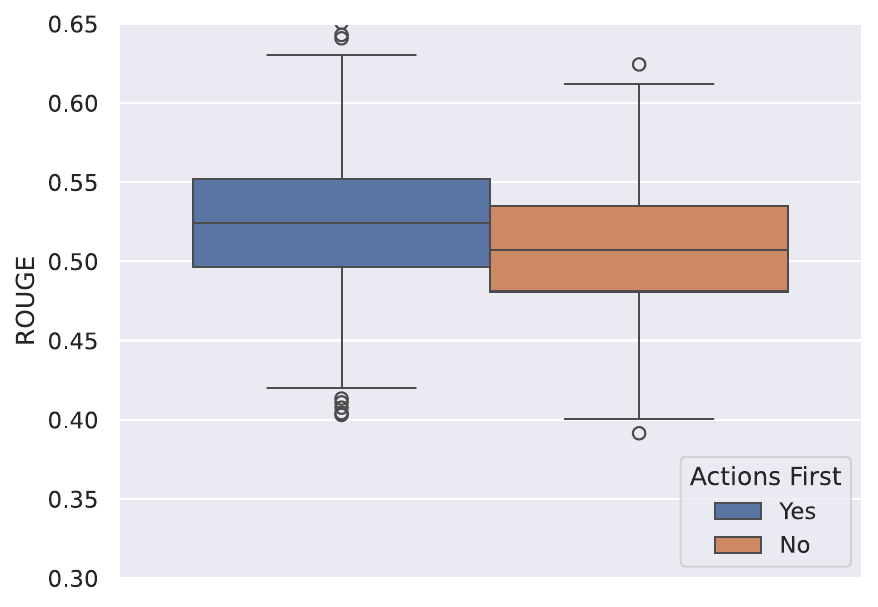}}
    \label{fig:rouge_weigh_cinfigs}
    \hfil
    \subfloat[Effects on action quality.]{\includegraphics[width=.45\linewidth]{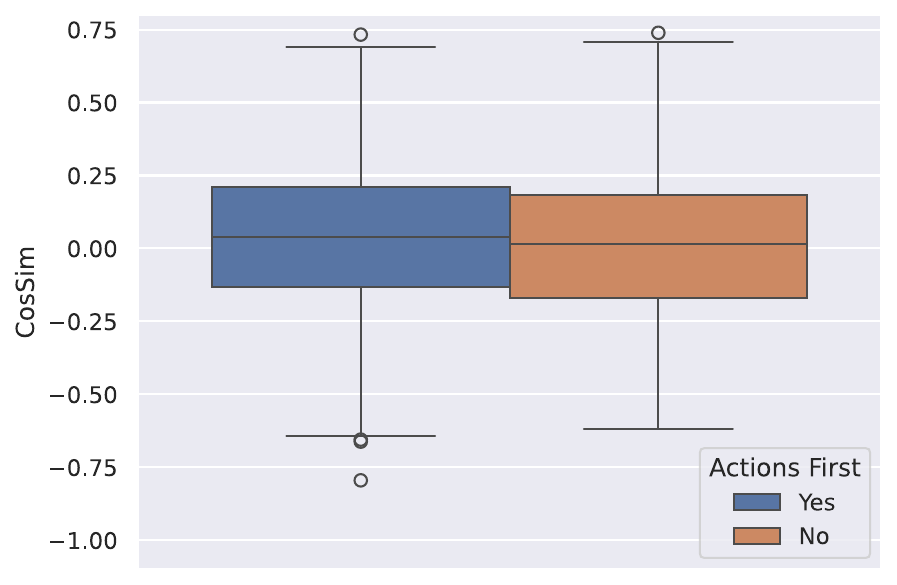}}
    \label{fig:cossim_weight_configs}
    \caption{Comparison between different orders of joint output: action tokens before or after the language statement.}
    \label{fig:actionsfirst}
\end{figure}
\subsection{How well does the model produce transparent statements?}\label{subsec:describe_actions}
\begin{figure*}
    \centering
    \includegraphics[width=\textwidth,trim={0 0.65cm 0 1.4cm},clip]{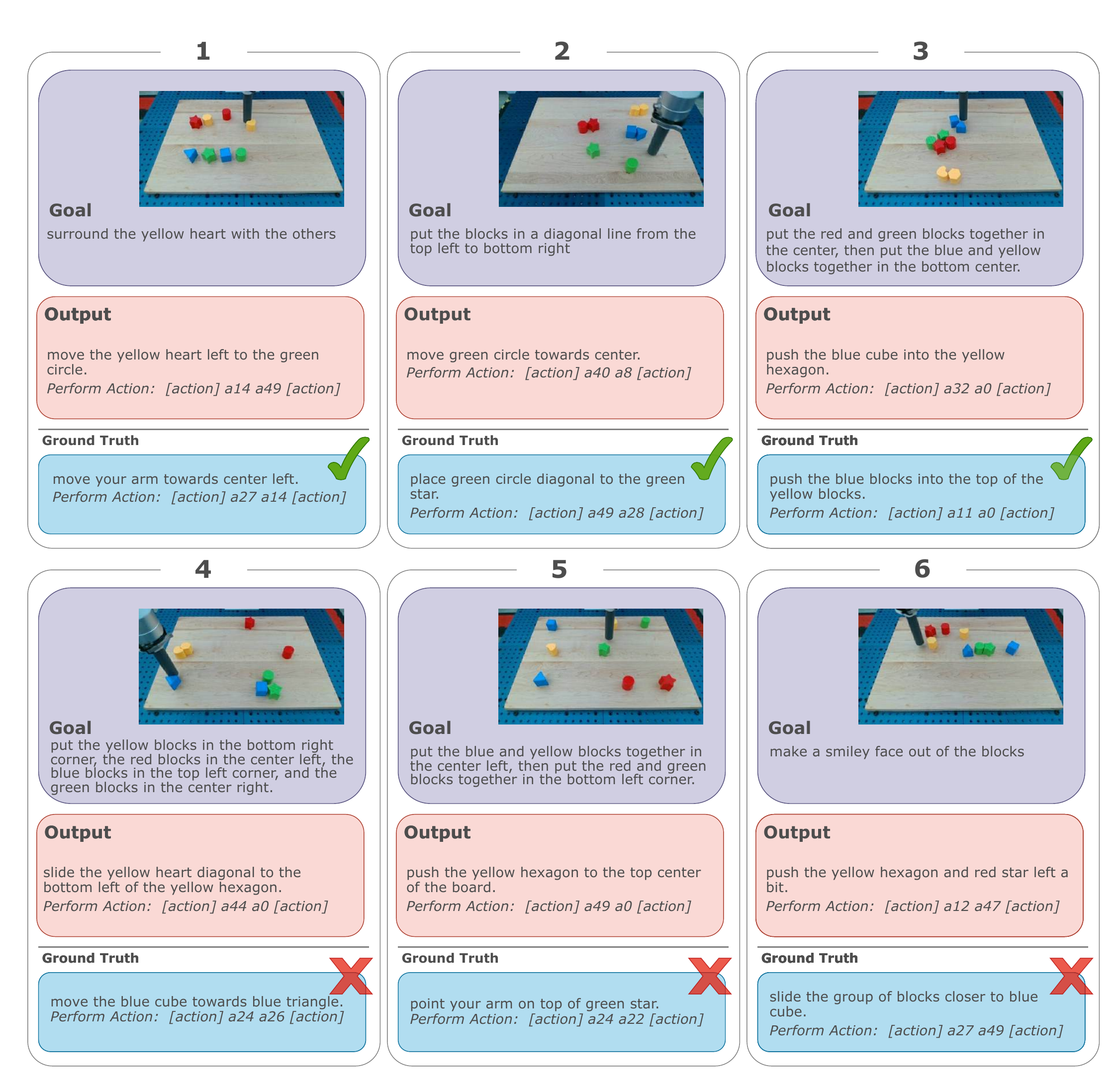}
    \caption{\textbf{Sample outputs} of our model on our test set including positive and negative samples. We removed the surrounding prompt-specific tokens for readability.}
    \label{fig:sample_output}
\end{figure*}

Based on the output of our test set, the model learns to generate comprehensive statements in natural language alongside a trajectory. 
Figure~\ref{fig:sample_output} presents a collection of sample input-output pairs.
We include three positive and three negative samples.
A sample is considered positive when the meaning of the generated transparent statement semantically resembles the ground truth statement or hints towards a similar action. 
The results further highlight the difficulty of evaluating learned transparent behavior. 
Even though the statement provided by the agent seems like a logical step towards approaching the provided goal, it can differ drastically from the ground truth wording. 
Sample 2 presents a good example of this, where the generated verb \textit{move} is semantically similar to the target \textit{place}.
Word and n-gram-based metrics, like BLEU and ROUGE, do not reflect this behavior, even though they can give an intuition, and can lead to lower scores.
In addition, the target direction of the pushing action is generated as \textit{towards center} while the ground truth refers to \textit{diagonal to the green star}.
Without supplementing visual input, it is also difficult to determine whether these two utterances refer to the same location.
When it comes to the negative samples, it can be challenging to determine whether the language output does present a viable option for achieving the goal despite the output not matching the ground truth.
% This further highlights the need for more advanced metrics to evaluate transparency via natural language in multimodal settings.
Regarding the actions, the predicted action tokens rarely exactly equal the ground truth tokens.
However, this does not mean that the generated trajectories are of low quality.
We refer to our other analyses which investigate the quality of trajectories after detokenizing the action tokens again.

% \subsection{Switching statements}
% The model's ability to predict the end of a sub-episode diminishes on novel scenarios in our test set, as seen in Table~\ref{tab:binary}. 
% We notice a drop of 25\% accuracy on the binary prediction on the test set result of 50\% compared to the 75\% on the validation set. 
% Showing that even though the model learns this prediction it is not generalizable and constitutes a difficult task.
% \begin{table}[ht]
%     \centering
%     \caption{Accuracy of determining when a statement is no longer valid and needs to be switched.}
%     \begin{tabular}{c|cc}
%          &  Validation & Test\\\hline
%         Accuracy & 75\% & 50\%\\ 
%     \end{tabular}
%     \label{tab:binary}
% \end{table}

\subsection{Action Before Language}
We find that producing the action tokens first and the statement last leads to more accurate action tokens than vice versa.
The results in Table~\ref{tab:action_first} show a slightly lower BLEU score but a higher ROUGE score when producing the language statement last. 
The increased ROUGE score means that this model generates more words that are part of the target statement than in the setting that produces the statement last.
However, it also generates more words that are not part of the ground truth, resulting in a lower BLEU score.
Figure~\ref{fig:actionsfirst} further highlights this observation.

%This could happen for example if your sys1 is outputting results which contain words from the references (upping the Rouge), but also many words which the references didn't include (lowering the Bleu). sys2, as it seems, is giving results for which most words outputted do appear in the human references (upping the Blue), but also missing many words from its results which do appear in the human references.

\begin{table}[ht]
    \caption{State Inclusion}
    \label{tab:state}
    \renewcommand{\arraystretch}{1.3}
    \centering
    \begin{tabular}{c|cccc}
        % State & CosSim$\uparrow$ & MSE$\downarrow$ & BLEU$\uparrow$ & ROUGE$\uparrow$ \\\hline
        % \ding{55}   & &  \\ 
        % \ding{51} &  &  \\ 
        
        % State &  ROUGE$\uparrow$ & BLEU$\uparrow$ & CosSim$\uparrow$ & MSE$\downarrow$ \\\hline
        % \ding{55} & 0.4463$\pm$0.04 & 0.1360$\pm$0.04 & \textbf{-0.0045$\pm$0.26} & 0.0024$\pm$0 \\ 
        % \ding{51} & \textbf{0.4596$\pm$0.04} & \textbf{0.1495$\pm$0.04} & -0.0196$\pm$0.26 & 0.0024$\pm$0 \\ 
        
        State &  ROUGE$\uparrow$ & BLEU$\uparrow$ & CosSim$\uparrow$ & MSE$\downarrow$ \\\hline
        \ding{55} & 0.4463 & 0.1360& \textbf{-0.0045} & 0.0024\\ 
        \ding{51} & \textbf{0.4596} & \textbf{0.1495} & -0.0196 & 0.0024 \\ 
    \end{tabular}
\end{table}

\subsection{Robot State Inclusion}
\begin{figure}[ht]
    \centering
    \subfloat[Effect on language modeling.]{\includegraphics[width=.45\linewidth]{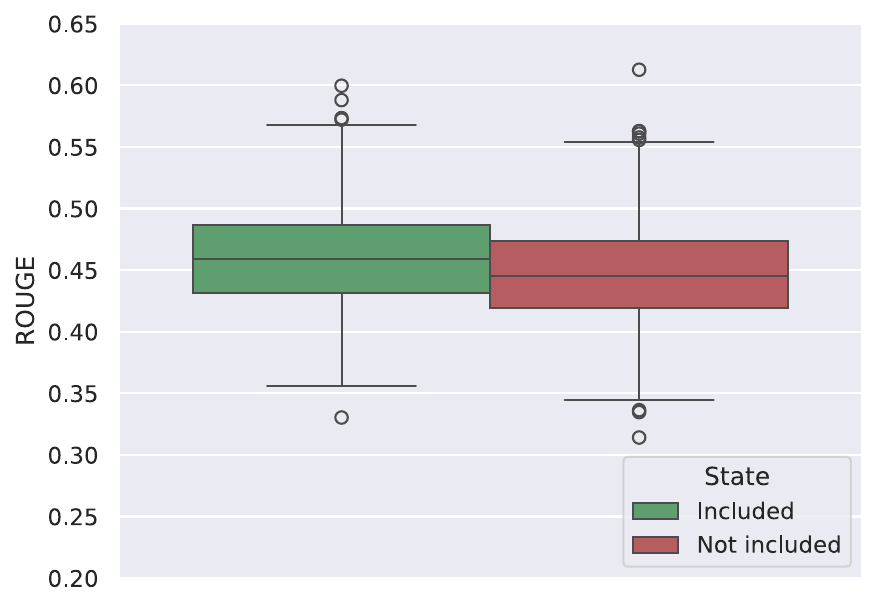}
    \label{fig:rouge_state_inclusion}}
    \hfil
    \subfloat[Effect on action quality.]{\includegraphics[width=.48\linewidth]{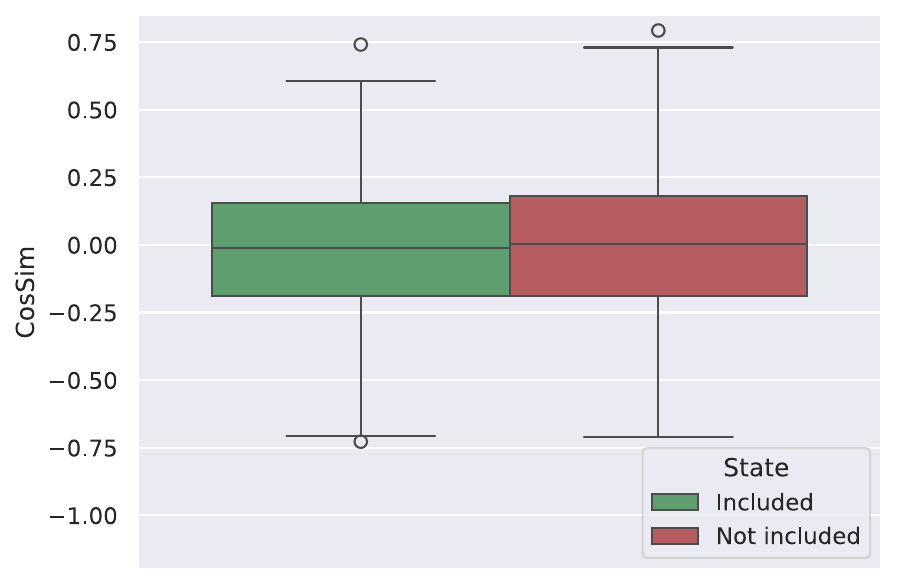}
    \label{fig:cossim_state_inclusion}}
    \caption{Effects of including the tokenized state vector in the input prompt.}
\end{figure}

From the results in Table~\ref{tab:state} and Figure~\ref{fig:rouge_state_inclusion} we find that the inclusion of state tokens in our input prompt has a slightly beneficial impact on the language generation capabilities of our model. 
The quality of the action remains the same, as can be seen in Figure~\ref{fig:cossim_state_inclusion}. 
It should be noted that introducing additional tokens for the robot state possibly increases the size of the word embedding, leading to increased computational cost.
\subsection{Tokenization Resolution}
\begin{figure}[ht]
    \centering
    \includegraphics[width=0.8\linewidth]{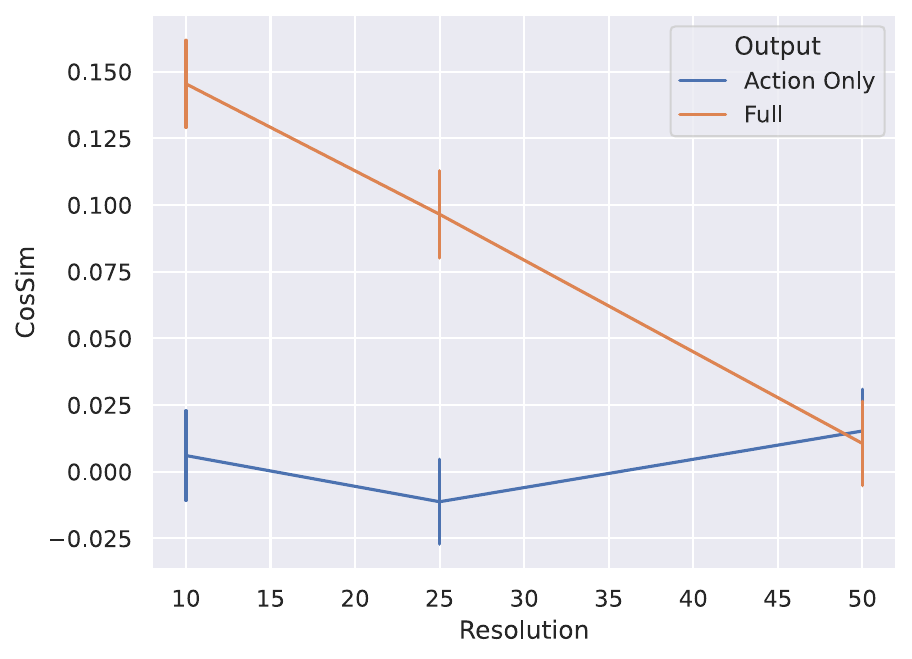}
    \caption{Effects of varying action tokenization resolutions on action quality.}
    \label{fig:cossim_resolution}
\end{figure}

We investigate the influence of different resolutions of the action tokens on the quality of the produced trajectories and language statements.
We hypothesize that a lower resolution (fewer discretization bins) results in higher scores.
We can observe this when producing the joint output of the language statement and the action trajectories, as shown in Figure~\ref{fig:cossim_resolution}.
When only producing the action tokens we do not observe this decreasing trend, and the action trajectory quality stays roughly within the same range.
The precise measurements can be found in Table~\ref{tab:resolution}.
In addition, generating both the transparent statement and the action trajectory can be observed to have a positive impact on the quality of the action trajectories.

\begin{table}[ht]
    \caption{Tokenization Resolution}
    \label{tab:resolution}
    \renewcommand{\arraystretch}{1.3}
    \centering
    \begin{tabular}{cc|cccc}        
        % Resolution & Output & ROUGE$\uparrow$ & BLEU$\uparrow$ & CosSim$\uparrow$ & MSE $\downarrow$ \\\hline
        % 10 & Action & - &  - &  0.0060$\pm$0.27 & 0.0024$\pm$0  \\ 
        % 25 & Action  &  - &  - & -0.0113$\pm$0.25 & 0.0023$\pm$0 \\  
        % 50 & Action &  - &  -  & 0.0152$\pm$0.25 & 0.0023$\pm$0 \\ 
        % 10 & Full & \textbf{0.5471$\pm$0.04} & \textbf{0.1696$\pm$0.04} & \textbf{0.1454$\pm$0.26} & \textbf{0.0021$\pm$0} \\ 
        % 25 & Full & 0.5338$\pm$0.04 & 0.1560$\pm$0.04 & 0.0965$\pm$0.26 & 0.0022$\pm$0 \\ 
        % 50 & Full & 0.5255$\pm$0.04 & 0.1511$\pm$0.04 & 0.0106$\pm$0.25 & 0.0024$\pm$0 \\ 
        
        % Resolution & Output & ROUGE$\uparrow$ & BLEU$\uparrow$ & CosSim$\uparrow$ & MSE $\downarrow$ \\\hline
        % 10 & Action & - &  - &  0.0060$\pm$0.27 & 0.0024$\pm$0  \\ 
        % 25 & Action  &  - &  - & -0.0113$\pm$0.25 & 0.0023$\pm$0 \\  
        % 50 & Action &  - &  -  & 0.0152$\pm$0.25 & 0.0023$\pm$0 \\ 
        % 10 & Full & \textbf{0.5471$\pm$0.04} & \textbf{0.1696$\pm$0.04} & \textbf{0.1454$\pm$0.26} & \textbf{0.0021$\pm$0} \\ 
        % 25 & Full & 0.5338$\pm$0.04 & 0.1560$\pm$0.04 & 0.0965$\pm$0.26 & 0.0022$\pm$0 \\ 
        % 50 & Full & 0.5255$\pm$0.04 & 0.1511$\pm$0.04 & 0.0106$\pm$0.25 & 0.0024$\pm$0 \\ 
        
        Resolution & Output & ROUGE$\uparrow$ & BLEU$\uparrow$ & CosSim$\uparrow$ & MSE$\downarrow$ \\\hline
        10 & Action & - &  - &  0.0060 & 0.0024  \\ 
        25 & Action  &  - &  - & -0.0113 & 0.0023 \\  
        50 & Action &  - &  -  & 0.0152 & 0.0023 \\ 
        10 & Full & \textbf{0.5471} & \textbf{0.1696} & \textbf{0.1454} & \textbf{0.0021} \\ 
        25 & Full & 0.5338 & 0.1560 & 0.0965 & 0.0022 \\ 
        50 & Full & 0.5255 & 0.1511 & 0.0106 & 0.0024 \\ 
    \end{tabular}
\end{table}

\subsection{Pretraining Influence}
As illustrated in Table~\ref{tab:pretrain} pretraining our models on the RoboVQA dataset leads to a higher quality of action trajectories than without pretraining.
At the same time, we notice slightly higher scores in the language metrics. 
Note that when only producing the action tokens as output, the pretraining on a language generation task does not increase the accuracy of the action tokens.
A reason for this could be the fact that the task of action generation introduces new tokens to the vocabulary that have not been encountered during the pretraining phase, lowering its effectiveness.
\begin{table}[ht]
    \caption{Pretraining}
    \label{tab:pretrain}
    \renewcommand{\arraystretch}{1.3}
    \centering
    \begin{tabular}{cc|cccc}
        % Checkpoint & Output & ROUGE$\uparrow$ &  BLEU$\uparrow$ & CosSim$\uparrow$ & MSE$\downarrow$ \\\hline
        % None & Action & - & - & 0.0152$\pm$0.25 & 0.0023$\pm$0 \\ 
        % None & Full & 0.5255$\pm$0.04 & 0.1511$\pm$0.04 & 0.0106$\pm$0.25 & 0.0024$\pm$0 \\ 
        % RoboVQA & Action & - & -  & 0.0108$\pm$0.25 & 0.0023$\pm$0 \\ 
        % RoboVQA & Full & \textbf{0.5309$\pm$0.04} & \textbf{0.1541$\pm$0.04} & \textbf{0.1601$\pm$0.24} & \textbf{0.0020$\pm$0} \\ 
        % Checkpoint & Output & ROUGE$\uparrow$ &  BLEU$\uparrow$ & CosSim$\uparrow$ & MSE$\downarrow$ \\\hline
        % None & Action & - & - & 0.0152$\pm$0.25 & 0.0023$\pm$0 \\ 
        % None & Full & 0.5255$\pm$0.04 & 0.1511$\pm$0.04 & 0.0106$\pm$0.25 & 0.0024$\pm$0 \\ 
        % RoboVQA & Action & - & -  & 0.0108$\pm$0.25 & 0.0023$\pm$0 \\ 
        % RoboVQA & Full & \textbf{0.5309$\pm$0.04} & \textbf{0.1541$\pm$0.04} & \textbf{0.1601$\pm$0.24} & \textbf{0.0020$\pm$0} \\ 
        
        Checkpoint & Output & ROUGE$\uparrow$ &  BLEU$\uparrow$ & CosSim$\uparrow$ & MSE$\downarrow$ \\\hline
        None & Action & - & - & 0.0152 & 0.0023 \\ 
        None & Full & 0.5255 & 0.1511 & 0.0106 & 0.0024 \\ 
        RoboVQA & Action & - & -  & 0.0108 & 0.0023 \\ 
        RoboVQA & Full & \textbf{0.5309} & \textbf{0.1541} & \textbf{0.1601} & \textbf{0.0020} \\ 
    \end{tabular}
\end{table}

\subsection{Freezing the Vision Encoder}
We observe little differences on the language generation performance ($\pm$0.04 on the ROUGE and BLEU scores) when training only the LLM part of the model compared to the full model, as visible in Table~\ref{tab:action_prediction_performance}.
The performance of generating action trajectories increases slightly when training the full model.
It is likely that the model learns to pay more attention to different novel visual features when having to generate the next trajectory, resulting in this behavior.
\begin{table}[ht]
    \caption{Results Weight and Output Configuration}
    \label{tab:action_prediction_performance}
    \renewcommand{\arraystretch}{1.3}
    \centering
    \begin{tabular}{cc|cccc}
        %  Training & Output & CosSim$\uparrow$ & MSE$\downarrow$ & BLEU$\uparrow$ & ROUGE$\uparrow$ \\\hline
        % LLM & Action & -0.0069$\pm$0.26 & 0.0023$\pm$0 & -  \\ 
        % LLM & Full & -0.0044$\pm$0.26 & 0.0024$\pm$0 &   \\ 
        % Full & Action & \textbf{0.0121$\pm$0.25} & 0.0023$\pm$0 & - \\ 
        % Full & Full & -0.0018$\pm$0.24 & 0.0024$\pm$0 &  &  \\ 

         % Training & Output & ROUGE$\uparrow$ & BLEU$\uparrow$ & CosSim$\uparrow$ & MSE$\downarrow$ \\\hline
         % LLM & Action & - & - & -0.0069$\pm$0.26 & 0.0023$\pm$0 \\
         % LLM & Full & \textbf{0.4462$\pm$0.04} & 0.1360$\pm$0.04 & -0.0044$\pm$0.26 & 0.0024$\pm$0 \\
         % Full & Action & - & - & \textbf{0.0121$\pm$0.25} & 0.0023$\pm$0 \\
         % Full & Full & 0.4436$\pm$0.04 & \textbf{0.1375$\pm$0.04} & -0.0018$\pm$0.24 & 0.0024$\pm$0 \\
         
         Training & Output & ROUGE$\uparrow$ & BLEU$\uparrow$ & CosSim$\uparrow$ & MSE$\downarrow$ \\\hline
         LLM & Action & - & - & -0.0069 & 0.0023\\
         LLM & Full & \textbf{0.4462} & 0.1360 & -0.0044 & 0.0024 \\
         Full & Action & - & - & \textbf{0.0121} & 0.0023 \\
         Full & Full & 0.4436 & \textbf{0.1375} & -0.0018& 0.0024 \\
    \end{tabular}
\end{table}

\section{Discussion}\label{sec:discussion}
Learning transparency is an inherently difficult task and depends on many factors. 
In this work, we opt for learning natural language statements that were provided by large-scale annotations. 
While learning by imitating in this way is certainly an option, datasets with ground truth language annotations for transparent statements in robotics are sparse and expensive to create, calling for novel methods to train transparent policies.
Even when these annotations are given, creating both language utterances and actions does not mean that these two outputs are semantically aligned; e.g., there is no guarantee that the description adheres to the executed trajectory.
Future work could address this by explicitly investigating measures that evaluate the quality and alignment of actions and language.
With respect to the semantic differences between generated and ground truth language, our results further highlight in Section~\ref{subsec:describe_actions} the need for more advanced metrics to evaluate transparency via natural language.
In addition, transparency should incorporate the human's perspective~\cite{Weller2019TransparencyMA}. 
What one human might perceive as transparent could not have the same effect on another. %TODO: cite
Further research could investigate incorporating the human as context to provide different statements depending on the user's preference.

A problem we faced when training the model to predict an action embedded into the language output using the traditional Cross-Entropy Loss is that this loss assigns the same importance to every token, which leads to the model not differentiating between different degrees of error when generating the action tokens. 
For example, if the goal action was "a20 a25", the corresponding error will not necessarily reflect the trajectory deviation for the output of "a5 a45" (large difference in trajectory) and the output of "a21 a24" (similar trajectory).
Addressing this could prove highly useful in future work using VLMs for policy generation.

Regarding a practical implementation, it is not necessary to provide a new transparent statement at each timestep, but rather once the described action has been performed.
Such a mechanism could be implemented with a specific prompt or an external module which only asks for a new transparent statement when certain conditions are met.
% To model this change in statements, we provide the model with another prompt asking whether it should update the verbal statement in the next time step. 
% We utilize the duration of the sub-episodes as the ground truth.
% Thus, the model should learn to predict the end of a sub-episode once the last low-level action was performed.
% If so, we can query the model for a new action description and only query actions until a new description is required.

\section{Conclusion}\label{sec:conclusion}
Driven by the need for more transparent robots, we presented a method to train agents to be transparent about their behavior using natural language while simultaneously learning low-level action trajectories.
Our model learns transparent behavior alongside a policy by combining both tasks into a single supervised language generation problem.
While we show promising results on the Language-Table dataset and find that transparency can benefit the model's action quality, we also highlight a need for methods to semantically analyze the quality of transparent agents, which we leave for investigation in further research.

% \appendix
% \section{Training Configuration}
% \label{app:config}

% \section{Prompt templates}
% \label{app:prompts}
% \begin{table}[ht]
%     \centering
%     \begin{tabular}{c|c|c}
%         Model & Question Template & Answer Template \\\hline
%         PaliGemma & TBD & TBD \\
%         BLIP-2 & TBD & TBD \\
%     \end{tabular}
%     \caption{Caption}
%     \label{tab:my_label}
% \end{table}

% \section{Pretraining Results}
% \label{app:pretraining}

% \end{document}

\bibliographystyle{IEEEtran}
\bibliography{Embodied_Agents}

\end{document}